\documentclass{article}

\usepackage[utf8]{inputenc}
\usepackage[margin=1in]{geometry} 
\usepackage[T1]{fontenc}
\usepackage{amsmath}
\usepackage{amssymb}
\usepackage{amsfonts}
\usepackage{graphicx}
\usepackage{grffile}
\usepackage{booktabs}
\usepackage{multirow}
\usepackage{natbib}
\usepackage{algorithm}
\usepackage{algpseudocode}     
\usepackage{xcolor}
\usepackage{microtype}
\usepackage{hyperref}
\usepackage{tabularx}
\usepackage{caption}
\usepackage{makecell}
\usepackage{adjustbox}
\usepackage{placeins}

\title{Traceable Cross-Source RAG for Chinese Tibetan Medicine Question Answering}
\author{Fengxian Chen, Zhilong Tao, Jiaxuan Li, Yunlong Li, Qingguo Zhou\thanks{Corresponding author. Email: \texttt{zhouqg@lzu.edu.cn}.}\\
School of Information Science \& Engineering, Lanzhou University\\
Lanzhou, China}

\begin{document}
\maketitle

\begin{abstract}
Retrieval-augmented generation (RAG) promises grounded question answering, yet domain settings with multiple heterogeneous knowledge bases (KBs) remain challenging. In Chinese Tibetan medicine, encyclopedia entries are often dense and easy to match, which can dominate retrieval even when classics or clinical papers provide more authoritative evidence.
We study a practical setting with three KBs (encyclopedia, classics, and clinical papers) and a 500-query benchmark (cutoff $K{=}5$) covering both single-KB and cross-KB questions. We propose two complementary methods to improve traceability, reduce hallucinations, and enable cross-KB verification. First, DAKS performs KB routing and budgeted retrieval to mitigate density-driven bias and to prioritize authoritative sources when appropriate. Second, we use an alignment graph to guide evidence fusion and coverage-aware packing, improving cross-KB evidence coverage without relying on naive concatenation. All answers are generated by a lightweight generator, \textsc{openPangu-Embedded-7B}. Experiments show consistent gains in routing quality and cross-KB evidence coverage, with the full system achieving the best CrossEv@5 while maintaining strong faithfulness and citation correctness.
\end{abstract}
\section{Introduction}
\label{sec:intro}

Large Language Models (LLMs) are increasingly used for information seeking and question answering. However, in high stakes domains, fluent responses are not enough. LLMs can hallucinate facts and fabricate evidence, which harms trust and safety \citep{huang2025hallucination_survey,sahoo2024hallucination_survey,yu2024auepora}. Retrieval Augmented Generation (RAG) mitigates this issue by grounding generation on external corpora.
RAG has shown clear gains in clinical tasks when retrieval uses curated resources \citep{zakka2024almanac,amugongo2025rag_healthcare_review}. Yet recent analyses and benchmarks show that RAG remains fragile. Errors come from retrieval failures, noisy context, and unfaithful use of evidence \citep{xiong-etal-2024-benchmarking,gao-etal-2023-enabling,es-etal-2024-ragas,yan2024crag,asai2024selfrag}.

Traditional medicine question answering has stronger requirements on provenance. In Chinese settings, Traditional Chinese Medicine (TCM) and Tibetan Medicine (TM) knowledge is scattered across heterogeneous sources.
These sources include classical canons, modern clinical papers, and encyclopedic summaries. They differ in writing style, term usage, and authority. Recent work has built TCM and TM question answering systems with knowledge graphs (KGs) or LLM-based assistants \citep{li2024dcmed_kgqa,qin2025ragcpmf,qin2025tcmlcm,dong2021tibetan_kgqa,pei2024tibetan_materia_medica_kg}. In parallel, domain benchmarks and models for Chinese medical NLP and TCM evaluation are emerging \citep{zhang2022cblue,wang2023huatuo,yu2024tcmd,wang2024domainrag}. However, existing systems typically index a single source or merge all sources into one flat corpus. They rarely study how source heterogeneity affects retrieval choices, evidence fusion, and traceability. This leaves a clear gap for TM question answering in Chinese, where multi-source partitioned knowledge bases are common.

This paper targets a practical but underexplored setting. We assume TM knowledge is split into separate knowledge bases by category, such as classics, clinical literature, and encyclopedia. Our goal is to make RAG answers more traceable, reduce hallucinations, and support cross-source verification. Two challenges are central.
First, dense sources can dominate retrieval. In our data, encyclopedia passages are short and information dense, which makes retrievers prefer them. But for many clinical or doctrine sensitive questions, encyclopedic summaries are not the best evidence. This mismatch can lead to plausible but poorly grounded answers. This problem is related to resource selection in federated search, but the objective here also involves authority and provenance \citep{wang2024feb4rag,wang2024resllm}.

Second, naive cross source fusion can reduce precision.
A common approach is to concatenate retrieved passages from multiple sources and let the LLM decide.
But long context models can be sensitive to passage order and position, which makes such fusion unstable
\citep{liu-etal-2024-lost-middle}.
Hybrid retrieval methods and KG-guided reranking are promising, but current work mainly evaluates general domains
\citep{yu2022kgfid,sarmah2024hybridrag}.
In TM, the problem is amplified by the semantic gap between classical terminology and modern clinical writing.
As a result, simply adding more passages often increases noise and harms evidence attribution.

We present a RAG framework designed for traceable TM question answering in Chinese under partitioned multi source knowledge bases.
Our approach is evaluated with a dataset and metrics that explicitly measure correctness and traceability, beyond surface fluency
\citep{gao-etal-2023-enabling,es-etal-2024-ragas,xiong-etal-2024-benchmarking,yu2024auepora}.
Our contributions are:
\begin{itemize}
    \item We formalize traceable cross source TM question answering in Chinese, where knowledge is split into heterogeneous knowledge bases by category.
    \item We propose two complementary components that address one goal: source aware retrieval to avoid density driven source bias, and KG-based cross source alignment to guide evidence fusion and reranking for better precision and attribution.
    \item We design an evaluation protocol with fine grained measurements of answer correctness, evidence support, and cross source verification, and we report strong improvements over competitive RAG baselines.
\end{itemize}

\section{Related Work}

\paragraph{Attribution and evaluation for retrieval-augmented generation.}
A central motivation for retrieval-augmented generation (RAG) is to improve factuality by grounding answers in external evidence, yet reliable attribution remains non-trivial.
ALCE formalizes citation-aware generation and proposes automatic metrics to assess citation quality beyond answer fluency and correctness \citep{gao-etal-2023-enabling}.
Complementarily, RAGAs provides a reference-free evaluation toolkit that decomposes RAG quality into retrieval and generation dimensions, enabling rapid iteration without expensive human labels \citep{es-etal-2024-ragas}.
For domain settings, MIRAGE benchmarks medical RAG across many configurations and highlights that even strong pipelines can remain sensitive to component choices \citep{xiong-etal-2024-benchmarking}.
These efforts motivate evaluation protocols that jointly consider answer correctness, evidence support, and citation reliability, which are critical for high-stakes knowledge domains.

\paragraph{Evidence ordering and long-context sensitivity.}
When multiple pieces of evidence are concatenated into a single prompt, large language models may underuse relevant information depending on its position.
The ``lost-in-the-middle'' phenomenon shows that performance can degrade when key evidence is placed in the middle of long contexts, even for long-context models \citep{liu-etal-2024-lost-middle}.
This observation is especially relevant for multi-source RAG, where naive concatenation or arbitrary ordering of retrieved passages can reduce precision.
It motivates evidence organization strategies that are robust to ordering effects and can better support traceable, evidence-backed answers.

\paragraph{Graph-augmented retrieval and traditional medicine question answering.}
Knowledge graphs (KGs) are often introduced into RAG to encode structured relations among entities and to improve evidence selection.
HybridRAG combines vector retrieval with KG-based retrieval and reports improved retrieval and generation quality compared with using either signal alone \citep{sarmah-etal-2024-hybridrag}.
In traditional medicine, recent work has started to build RAG-style question answering systems for Traditional Chinese Medicine (TCM), typically by constructing a single vectorized corpus from curated textbooks and classical materials \citep{zhang-etal-2024-tcm-rag-llm}.
For Tibetan medicine, existing published systems more commonly focus on KG construction and KG-based question answering rather than multi-source RAG over partitioned corpora \citep{pei2024tibetan_materia_medica_kg,dong2021tibetan_kgqa}.
Overall, prior work provides useful building blocks (citation evaluation, RAG assessment, long-context analysis, and KG-enhanced retrieval), but it leaves a gap in studying \emph{partitioned, heterogeneous} traditional medicine knowledge bases where provenance, authority, and cross-library evidence alignment are first-order requirements.

\section{Method}
\label{sec:method}

\subsection{Problem Setup and Notation}
We consider three Chinese Tibetan-medicine knowledge bases (KBs), $\mathcal{K}=\{\mathsf{E},\mathsf{T},\mathsf{P}\}$, for encyclopedia, classics, and clinical papers.
Each KB $k$ is segmented into a set of chunks $\mathcal{D}_k=\{c\}$. Each chunk $c$ has text $x_c$ and metadata $m_c$ including KB id, document id, and a structural path. Given a query $q$, the system outputs an answer $\hat{a}$ and a cited evidence list $\mathcal{E}(q)=\langle c_1,\dots,c_K\rangle$ with stable chunk ids.

We assume two scorers: (i) a semantic retriever producing a relevance score $s_{\mathrm{ret}}(q,c)$ and (ii) an optional semantic ranker producing $s_{\mathrm{rank}}(q,c)$. We do not bind them to specific model families in the method section.

For cross-KB alignment, each chunk is associated with a typed entity set $E(c)$. We use a fixed type inventory (e.g., \textsc{Disease}, \textsc{Symptom}, \textsc{Drug}, \textsc{Formula}, etc.). We also extract a typed entity set $E(q)$ from the query.

\begin{figure}[htbp]
  \centering
  \includegraphics[width=\textwidth]{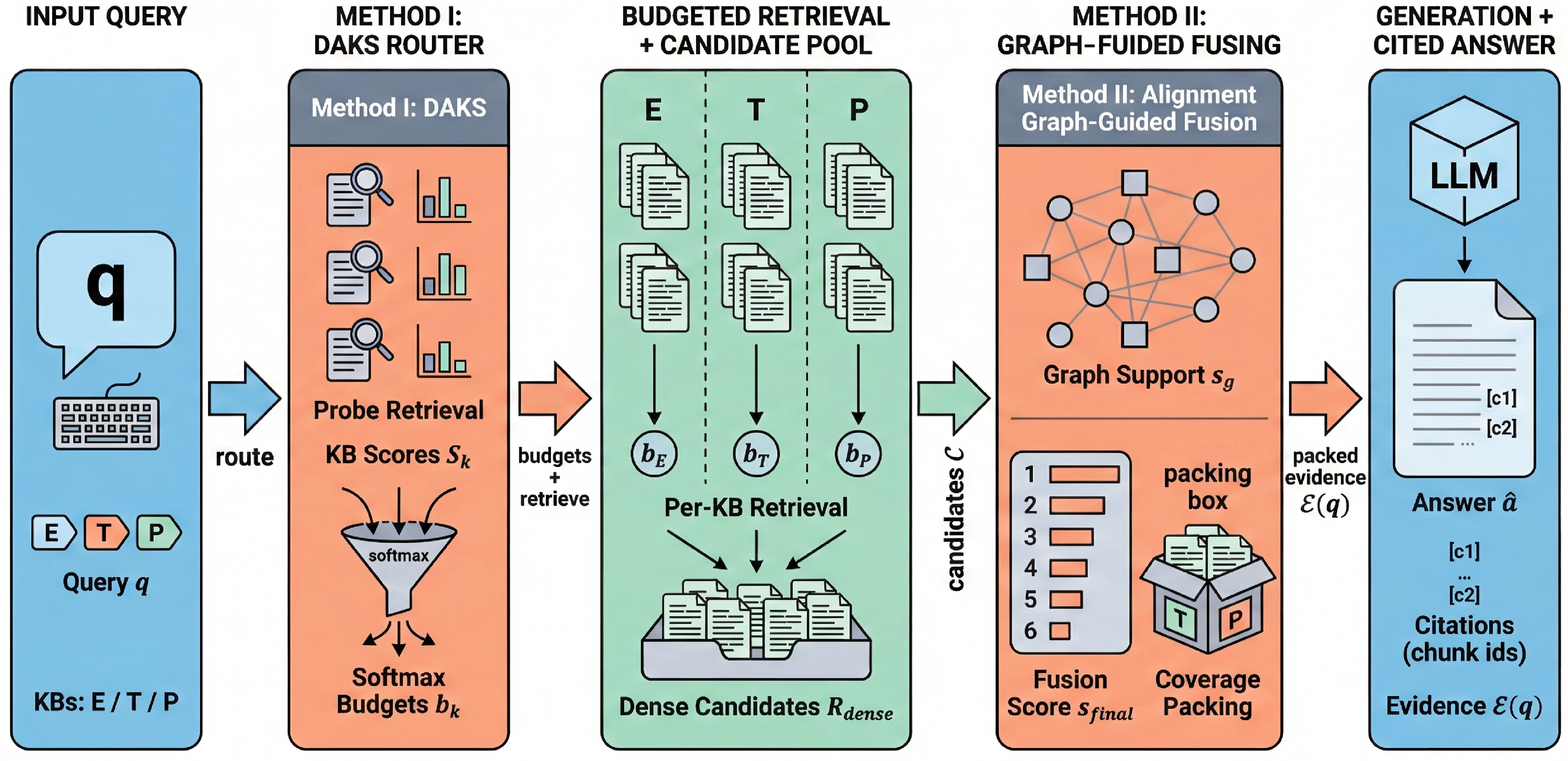}
  \caption{Overall system overview.}
  \label{fig:pipeline}
\end{figure}

Figure~\ref{fig:pipeline} gives an overview of our end-to-end pipeline, highlighting where Method~I (DAKS routing) and Method~II (graph-guided fusion) intervene.

\subsection{Method I: DAKS Routing with Budgeted Retrieval}
\label{sec:daks}
In our setting, KBs are heterogeneous. Encyclopedia chunks are short and information-dense, which often yields higher retrieval scores. This can bias retrieval toward $\mathsf{E}$ even when $\mathsf{T}$ or $\mathsf{P}$ is the preferred source. We therefore treat routing as a \emph{budgeted resource selection} problem.

\subsubsection{Probe retrieval and KB-level statistics}
For each KB $k$, we run a small probe retrieval to obtain $L$ candidates:
\begin{equation}
\mathcal{P}_k(q)=\mathrm{TopL}\big(\{(c, s_{\mathrm{ret}}(q,c)) : c\in \mathcal{D}_k\}\big).
\end{equation}
Let the sorted probe scores be $s^{(1)}_k \ge \cdots \ge s^{(L)}_k$. We compute a KB feature vector $\phi_k(q)$ to characterize the score distribution:
\begin{align}
\phi_k(q) = [&\underbrace{s^{(1)}_k}_{\text{peak}},
\underbrace{\frac{1}{M}\sum_{i=1}^{M}s^{(i)}_k}_{\text{top-}M\text{ mean}},
\underbrace{s^{(1)}_k - s^{(M)}_k}_{\text{margin}}, \\
&\underbrace{H(\pi_k)}_{\text{concentration}},
\underbrace{\mathrm{cov}_k(q)}_{\text{coverage}} ].
\end{align}
Here $\pi_k=\mathrm{Softmax}([s^{(1)}_k,\dots,s^{(L)}_k])$ and $H(\cdot)$ is entropy. A lower entropy indicates more concentrated high-score evidence.
$\mathrm{cov}_k(q)$ is a lightweight coverage proxy (e.g., distinct documents or sections in $\mathcal{P}_k(q)$), to avoid allocating all budget to a single redundant source.

We summarize DAKS routing and budgeted retrieval in Algorithm~\ref{alg:daks}.

\begin{algorithm}[t]
\caption{DAKS Routing with Budgeted Retrieval}
\label{alg:daks}
\begin{algorithmic}[1]
\Require query $q$; KBs $\mathcal{K}=\{\mathsf{E},\mathsf{T},\mathsf{P}\}$; total budget $B$; minimum budget $b_{\min}$; probe size $L$
\Require retriever score function $s_{\mathrm{ret}}(q,c)$; authority prior $a_k$
\Ensure KB scores $\{S_k\}$; KB ranking $\pi_{\mathrm{KB}}$; budgets $\{b_k\}$; dense candidate pool $\mathcal{R}_{\mathrm{dense}}$

\For{each KB $k \in \mathcal{K}$}
    \State $\mathcal{P}_k \gets \mathrm{TopL}\big(\{(c,s_{\mathrm{ret}}(q,c)) : c \in \mathcal{D}_k\}\big)$ \Comment{probe retrieval}
    \State Let sorted probe scores be $s_k^{(1)} \ge \cdots \ge s_k^{(L)}$
    \State $\pi_k \gets \mathrm{Softmax}([s_k^{(1)},\ldots,s_k^{(L)}])$
    \State $\phi_k(q) \gets [\text{peak } s_k^{(1)},\ \text{top-}M \text{ mean},\ \text{margin } (s_k^{(1)}{-}s_k^{(M)}),\ H(\pi_k),\ \mathrm{cov}_k(q)]$
    \State $S_k \gets \mathbf{w}^\top \phi_k(q) + \lambda a_k$
\EndFor

\State $\pi_{\mathrm{KB}} \gets \mathrm{SortDescending}(\{(k,S_k)\}_{k\in\mathcal{K}})$
\State $p_k \gets \mathrm{Softmax}(\{S_k\}_{k\in\mathcal{K}})$ for all $k$
\State $b_k \gets b_{\min} + \mathrm{Round}\big((B-|\mathcal{K}|b_{\min}) \cdot p_k\big)$ for all $k$

\State \textbf{Adjust budgets to sum to $B$:}
\While{$\sum_k b_k \neq B$}
    \If{$\sum_k b_k > B$}
        \State decrease $b_{k^\star}$ by 1 where $k^\star=\arg\max_k b_k$ and $b_{k^\star}>b_{\min}$
    \Else
        \State increase $b_{k^\star}$ by 1 where $k^\star=\arg\max_k p_k$
    \EndIf
\EndWhile

\State $\mathcal{R}_{\mathrm{dense}} \gets \emptyset$
\For{each KB $k \in \mathcal{K}$}
    \State $\mathcal{R}_{\mathrm{dense}} \gets \mathcal{R}_{\mathrm{dense}} \cup \mathrm{Top}\,b_k\big(\{(c,s_{\mathrm{ret}}(q,c)) : c \in \mathcal{D}_k\}\big)$
\EndFor
\State \Return $\{S_k\}, \pi_{\mathrm{KB}}, \{b_k\}, \mathcal{R}_{\mathrm{dense}}$
\end{algorithmic}
\end{algorithm}

\begin{figure}[t]
  \centering
  \includegraphics[width=\columnwidth]{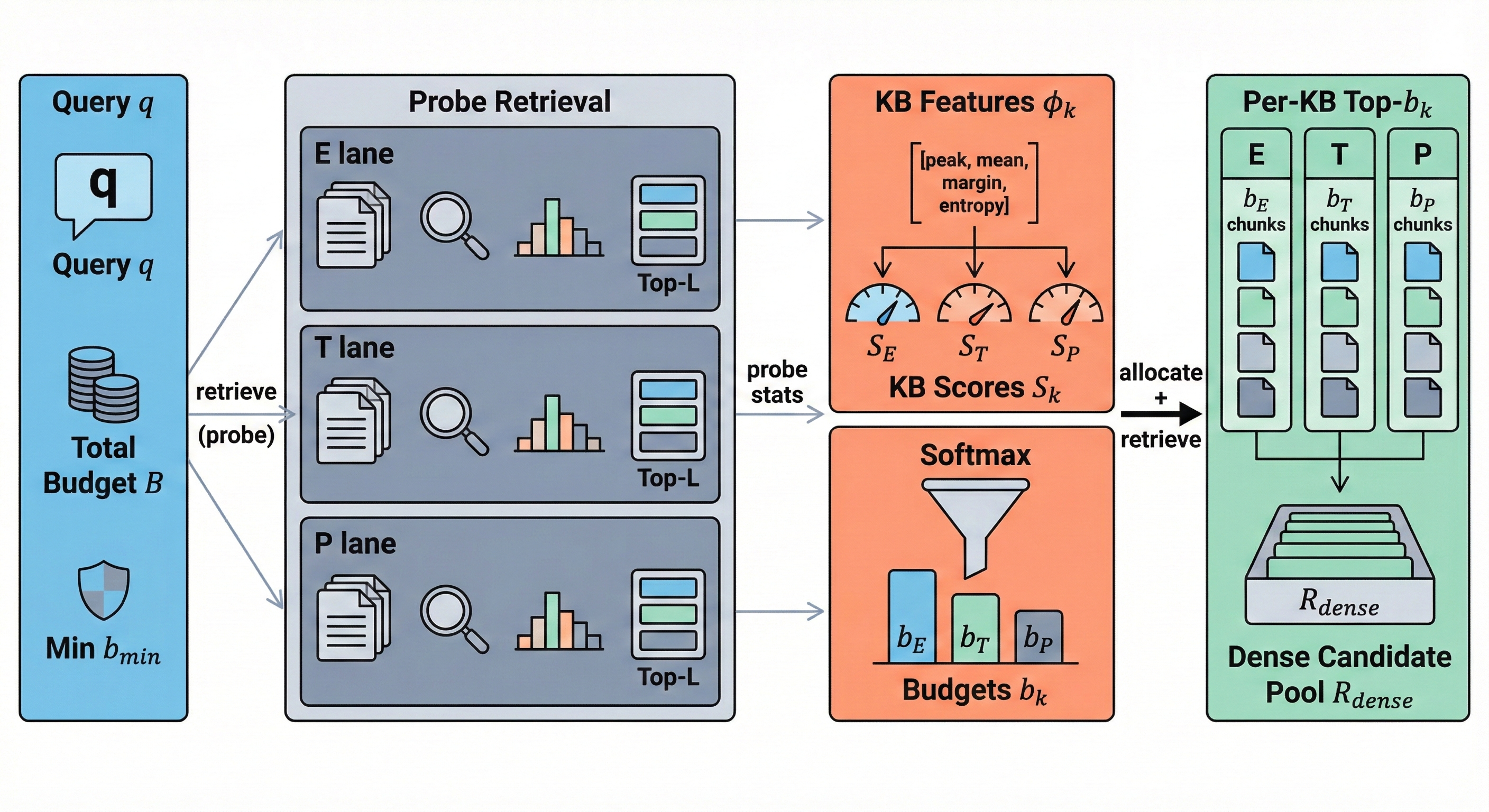}
  \caption{DAKS. We run lightweight probe retrieval in each KB, summarize score distributions into KB-level features, compute KB scores, and allocate a soft budget to form a balanced candidate pool.}
  \label{fig:daks}
\end{figure}

Figure~\ref{fig:daks} illustrates how DAKS uses probe statistics to score KBs and allocate per-KB budgets, mitigating density-driven encyclopedia dominance.

\subsubsection{Authority-aware KB scoring}
We add an authority prior $a_k$ to encode our preference over evidence provenance (e.g., $\mathsf{P}$ and $\mathsf{T}$ are often more authoritative than $\mathsf{E}$ for clinical queries).
We score each KB:
\begin{equation}
S_k(q)=\mathbf{w}^\top \phi_k(q) + \lambda a_k.
\end{equation}
This is a \emph{KB-level} relevance estimate that is not tied to any single chunk.

\subsubsection{Soft budget allocation and candidate pool}
Given a total retrieval budget $B$ and a minimum guarantee $b_{\min}$ for each KB, we allocate per-KB budgets using the KB scores:
\begin{equation}
b_k(q)=b_{\min} + \mathrm{Round}\big((B-|\mathcal{K}|b_{\min})\cdot p_k(q)\big),
\quad p_k(q)=\mathrm{Softmax}(S(q))_k.
\end{equation}
We then retrieve $b_k(q)$ candidates from each KB and merge them:
\begin{equation}
\mathcal{R}_{\mathrm{dense}}(q)=\bigcup_{k\in\mathcal{K}} \mathrm{Top}\,b_k(q)\big(\{(c,s_{\mathrm{ret}}(q,c)):c\in\mathcal{D}_k\}\big).
\end{equation}
We also keep the KB ranking induced by $S_k(q)$ for routing evaluation (primary-source and Top-2).

\subsection{Candidate Consolidation}
\label{sec:consolidation}
$\mathcal{R}_{\mathrm{dense}}(q)$ may contain fragmented evidence from long structured documents. We consolidate candidates into a cleaner list $\mathcal{C}(q)$ before cross-KB fusion.
First, we apply \emph{structure-aware expansion}. For each candidate chunk $c$, we add its local neighbors in the same document based on the structural path:
\begin{equation}
\mathrm{Expand}(c)=\{c\}\cup \mathrm{Nbr}(c),
\end{equation}
where $\mathrm{Nbr}(c)$ can include parent-section summaries or adjacent chunks within the same section.

Second, we deduplicate and enforce diversity. We remove near-duplicates and cap repeated chunks from a single document. This prevents one dense document from dominating the evidence list.

Finally, we compute a base score $s_{\mathrm{base}}(q,c)$ for all $c\in\mathcal{C}(q)$, which can combine retrieval relevance and optional ranking:
\begin{equation}
s_{\mathrm{base}}(q,c)=\mu \,\hat{s}_{\mathrm{ret}}(q,c) + (1-\mu)\,\hat{s}_{\mathrm{rank}}(q,c),
\end{equation}
where $\hat{\cdot}$ denotes score normalization within $\mathcal{C}(q)$. 

\subsection{Method II: Alignment Graph-Guided Fusion for Cross-KB Verification}
\label{sec:graphfusion}
Naive concatenation of multi-KB contexts can introduce noise and lower precision. It is also sensitive to evidence order in long prompts. Long-context studies show that relevant information placed in the middle can be underused, which makes evidence ordering and packing critical. We address this with an \emph{alignment graph} and a \emph{constrained evidence packing} objective.

\subsubsection{Alignment graph construction}
We build a bipartite graph $G=(V_C \cup V_E, \mathcal{E}_G)$. Chunk nodes $V_C$ correspond to chunks in all KBs, and entity nodes $V_E$ correspond to typed entities.
We add an edge $(c,e)$ if entity $e\in E(c)$. Each chunk node keeps its KB label $k(c)\in\mathcal{K}$.

\subsubsection{Graph-based bridge retrieval (optional)}
Given query entities $E(q)$ and consolidated candidates $\mathcal{C}(q)$, we collect seed entities from the top candidates:
\begin{equation}
E_{\mathrm{seed}}(q)=E(q)\cup \bigcup_{c\in \mathrm{TopS}(\mathcal{C}(q), s_{\mathrm{base}})} E(c).
\end{equation}
We then traverse $G$ from $E_{\mathrm{seed}}(q)$ for at most $h$ hops (entity--chunk alternation), and retrieve additional chunks:
\begin{equation}
\mathcal{R}_{\mathrm{graph}}(q)=\{c \in V_C : \exists e \in E_{\mathrm{seed}}(q), \ \mathrm{dist}_G(e,c)\le h\}.
\end{equation}
This step explicitly seeks cross-KB bridges (e.g., linking a classics concept to clinical evidence).

\begin{figure}[t]
  \centering
  \includegraphics[width=\columnwidth]{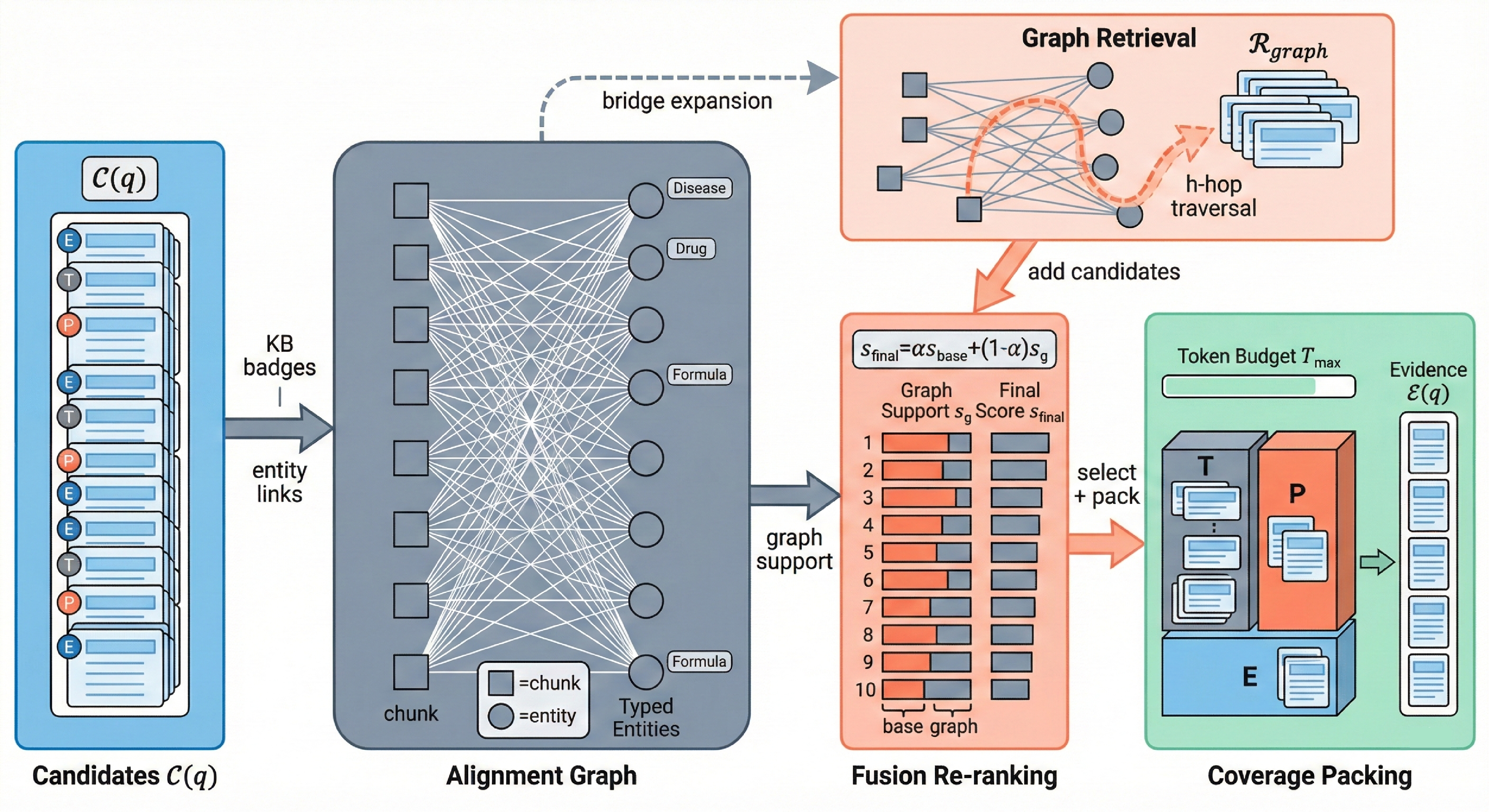}
  \caption{Alignment Graph-Guided Fusion. We build a chunk--entity alignment graph to compute graph support signals, fuse them with semantic relevance for reranking, and pack evidence under a token budget with cross-KB coverage constraints.}
  \label{fig:graphfusion}
\end{figure}

Figure~\ref{fig:graphfusion} visualizes our alignment graph and the fusion-and-packing procedure for cross-KB verification, from graph support scoring to coverage-aware evidence selection.

For any candidate chunk $c$ (from $\mathcal{C}(q)\cup \mathcal{R}_{\mathrm{graph}}(q)$), we compute a graph support score using two signals:
(1) entity overlap and (2) graph proximity to query/seed entities:
\begin{align}
o(q,c) &= |E(q)\cap E(c)|,\\
d(q,c) &= \min_{e\in E(c),\, e'\in E_{\mathrm{seed}}(q)} \mathrm{dist}_G(e,e').
\end{align}
We define:
\begin{equation}
s_g(q,c)=\eta_1 \log(1+o(q,c)) + \eta_2 \frac{1}{1+d(q,c)} + \eta_3 \,\mathbb{I}[k(c)\neq k_{\mathrm{major}}(q)],
\end{equation}
where $k_{\mathrm{major}}(q)$ is the top KB predicted by DAKS, and the last term encourages cross-KB verification. We fuse semantic relevance and graph support:
\begin{equation}
s_{\mathrm{final}}(q,c)=\alpha \hat{s}_{\mathrm{base}}(q,c) + (1-\alpha)\hat{s}_g(q,c).
\end{equation}
This yields a single ranked list $\mathcal{L}(q)$.

\subsubsection{Constrained evidence packing under coverage}
We select the final evidence list $\mathcal{E}(q)$ under a token budget $T_{\max}$. For each chunk $c$, let $\ell(c)$ be its token length. We solve a constrained selection problem:
\begin{align}
\max_{\mathcal{E}\subseteq \mathcal{L}(q)} \quad & \sum_{c\in \mathcal{E}} s_{\mathrm{final}}(q,c) \\
\text{s.t.}\quad & \sum_{c\in \mathcal{E}} \ell(c)\le T_{\max}, \\
& \forall k \in \mathcal{K}_{\mathrm{req}}(q),\ \exists c\in \mathcal{E} \ \text{with}\ k(c)=k,
\end{align}
where $\mathcal{K}_{\mathrm{req}}(q)$ is the required KB set (single-KB or multi-KB). We use a greedy procedure:
(i) first satisfy coverage by selecting the best chunk per required KB,
(ii) then fill remaining budget by descending $s_{\mathrm{final}}(q,c)$ with diversity caps.
This packing directly targets cross-KB verification and reduces failures caused by naive concatenation.

Algorithm~\ref{alg:pack} details our coverage-aware greedy evidence packing under a token budget.

\begin{algorithm}[t]
\caption{Coverage-aware Greedy Evidence Packing}
\label{alg:pack}
\begin{algorithmic}[1]
\Require ranked list $\mathcal{L}(q)$ with scores $s_{\mathrm{final}}(q,c)$; required KB set $\mathcal{K}_{\mathrm{req}}(q)$; token budget $T_{\max}$
\Require token length $\ell(c)$; optional doc cap $C_{\mathrm{doc}}$ (max chunks per doc)
\Ensure packed evidence list $\mathcal{E}(q)$

\State $\mathcal{E} \gets [\ ]$; $T \gets 0$; initialize doc counter map $\mathrm{cnt}(\cdot)\gets 0$
\State \textbf{Phase 1: satisfy coverage constraints}
\For{each required KB $k \in \mathcal{K}_{\mathrm{req}}(q)$}
    \State $c^\star \gets \arg\max_{c \in \mathcal{L}(q): k(c)=k}\ s_{\mathrm{final}}(q,c)$ s.t. $T+\ell(c)\le T_{\max}$
    \If{$c^\star$ exists}
        \State append $c^\star$ to $\mathcal{E}$; $T \gets T+\ell(c^\star)$; $\mathrm{cnt}(\mathrm{doc}(c^\star)) \gets \mathrm{cnt}(\mathrm{doc}(c^\star))+1$
    \EndIf
\EndFor

\State \textbf{Phase 2: fill remaining budget by descending score}
\For{each chunk $c$ in $\mathcal{L}(q)$ in order}
    \If{$c \notin \mathcal{E}$}
        \If{$T+\ell(c)\le T_{\max}$}
            \If{$\mathrm{cnt}(\mathrm{doc}(c)) < C_{\mathrm{doc}}$}
                \State append $c$ to $\mathcal{E}$; $T \gets T+\ell(c)$; $\mathrm{cnt}(\mathrm{doc}(c)) \gets \mathrm{cnt}(\mathrm{doc}(c))+1$
            \EndIf
        \EndIf
    \EndIf
\EndFor
\State \Return $\mathcal{E}(q)$
\end{algorithmic}
\end{algorithm}

\section{Experiments}
\label{sec:experiments}

\subsection{Dataset}
\label{sec:dataset}
We construct a Chinese Tibetan-medicine QA dataset over three KBs: encyclopedia ($\mathsf{E}$), classics ($\mathsf{T}$), and clinical papers ($\mathsf{P}$). Each instance contains a query, a short gold answer, and gold evidence at the chunk level (with KB labels, document metadata, and stable chunk ids). The dataset contains 500 queries and is balanced across four question types (25\% each): (i) definition, (ii) classics principles, (iii) clinical evidence, and (iv) cross-KB synthesis. For cross-KB synthesis, each example specifies a required KB set (at least two KBs) and a primary source KB.

Unlike typical supervised settings, we do not train any component on this dataset and do not create train/dev/test splits. All reported metrics are computed on the full 500-query set.

\subsection{Experimental Setup}
\label{sec:setup}

We index chunks from each KB independently. At inference time, we compute chunk relevance scores with a semantic retriever. We optionally apply a semantic reranker in the consolidation stage. All retriever and reranker parameters are kept fixed during evaluation.

We use \textsc{openPangu-Embedded-7B} as the answer generator for all methods \cite{chen2025pangu}. This choice highlights a lightweight, deployment-friendly LLM in a domain RAG setting. To ensure a fair comparison, we keep the same generation prompt and citation format across all methods.

Since we do not use a validation split, we set a single hyperparameter configuration and keep it fixed for
all experiments. This includes the probe size $L$, total retrieval budget $B$, minimum per-KB budget $b_{\min}$,
fusion weight $\alpha$ in Method~II, optional graph hop limit $h$, and token budget $T_{\max}$. We set the evaluation cutoff to $K{=}5$ for all @K metrics.

All automatic metrics, including RAGAS-style scores and citation checks, are computed by the same judge model, \textsc{GLM-4.7}, to avoid confounding effects from heterogeneous evaluators.

\subsection{Metrics}
\label{sec:metrics}
We evaluate at three levels with a consistent cutoff $K{=}5$.

\paragraph{KB routing (Method I).}
Primary-source accuracy (PrimaryAcc) and Top-2 hit (Top2Hit) are computed from the KB score ranking over
$\{\mathsf{E},\mathsf{T},\mathsf{P}\}$. We also report encyclopedia dominance rate (EDR), defined as the fraction of $\mathsf{E}$ chunks in the final evidence list.

\paragraph{Evidence selection and cross-KB verification (Method II).}
We report EvRecall@5 and EvNDCG@5 based on gold chunk ids. For cross-KB queries, we report CrossEv@5, which checks whether the top-5 evidence covers all required KBs. Unless otherwise stated, evidence-level metrics are computed over the cross-KB subset.

\paragraph{Answer quality, faithfulness, and citation correctness.}
We report reference-free RAG metrics (faithfulness, context precision/recall, and answer relevance) in the spirit of RAGAS. We also report citation-based checks inspired by ALCE to assess whether cited evidence supports the generated statements.

\subsection{Main Results}
\label{sec:mainresults}

Table~\ref{tab:main} summarizes end-to-end performance on all 500 queries. We compare against (i) single-KB retrieval baselines, (ii) pooled retrieval over merged KBs, and (iii) naive multi-KB concatenation. We further include two partial variants (\textsc{DAKS only} and \textsc{GraphFusion only}) to isolate the contributions of Method~I and Method~II.

Overall, \textsc{DAKS+GraphFusion} achieves the best cross-KB evidence coverage (CrossEv@5) while maintaining
strong faithfulness and citation correctness. We observe that enabling graph-guided fusion without DAKS routing can underperform on end-to-end answer relevance, despite strong evidence-level gains (Table~\ref{tab:fusion}), suggesting that graph-guided fusion benefits from a well-routed candidate pool produced by Method~I.

\begin{table}[htbp]
\centering
\small
\begin{adjustbox}{max width=\linewidth}
\begin{tabular}{lcccccc}
\toprule
Method & Faith. & CtxPrec & CtxRec & AnsRel & CrossEv@5 & CitCorr \\
\midrule
Single-KB ($\mathsf{E}$) & 0.654 & 0.415 & 0.834 & 0.902 & 0.720 & 0.756 \\
Single-KB ($\mathsf{T}$) & 0.837 & 0.223 & 0.734 & 0.833 & 0.750 & 0.791 \\
Single-KB ($\mathsf{P}$) & 0.810 & 0.245 & 0.705 & 0.805 & 0.680 & 0.680 \\
Merged KB & 0.785 & 0.195 & 0.750 & 0.795 & 0.650 & 0.720 \\
Naive Multi-KB Concat & 0.750 & 0.180 & 0.745 & 0.780 & 0.620 & 0.700 \\
DAKS only & 0.820 & 0.235 & 0.760 & 0.825 & 0.720 & 0.750 \\
GraphFusion only & 0.630 & 0.306 & 0.409 & 0.408 & 0.650 & 0.650 \\
\textbf{DAKS + GraphFusion (Full)} & 0.805 & 0.265 & 0.720 & 0.810 & 0.780 & 0.760 \\
\bottomrule
\end{tabular}
\end{adjustbox}
\caption{End-to-end performance on all 500 queries with cutoff $K{=}5$. All answers are generated by \textsc{openPangu-Embedded-7B}, and all automatic metrics are computed by \textsc{GLM-4.7}.}
\label{tab:main}
\end{table}
\FloatBarrier

\subsection{Routing Evaluation (Method I: DAKS)}
\label{sec:routing}

Table~\ref{tab:daks} evaluates KB routing quality.
DAKS improves primary-source prediction accuracy over uniform budgeting and merged-KB retrieval, and reduces encyclopedia dominance in the final evidence list, indicating its effectiveness in mitigating density-driven bias.

\begin{table}[htbp]
\centering
\small
\begin{adjustbox}{max width=\linewidth}
\begin{tabular}{lcccc}
\toprule
Method & PrimaryAcc & Top2Hit & EDR($\downarrow$) & AuthCov \\
\midrule
Uniform per-KB budget & 0.500 & 0.680 & 0.415 & 0.180 \\
Merged KB & 0.480 & 0.640 & 0.485 & 0.140 \\
DAKS (ours) & 0.560 & 0.700 & 0.362 & 0.230\\
\bottomrule
\end{tabular}
\end{adjustbox}
\caption{Routing evaluation with cutoff $K{=}5$. AuthCov measures whether evidence from authoritative KBs appears in top-5 (for relevant queries).}
\label{tab:daks}
\end{table}
\FloatBarrier

\subsection{Evidence Fusion Evaluation (Method II)}
\label{sec:fusion}

Table~\ref{tab:fusion} evaluates evidence ranking quality and cross-KB verification on the cross-KB subset. We compare naive concatenation, a score-only reranking baseline, and our graph-guided fusion. Graph-support fusion substantially improves CrossEv@5 and reduces encyclopedia dominance. Enabling graph bridge retrieval further improves recall and ranking quality, suggesting that the alignment graph
retrieves complementary evidence across KB boundaries.

\begin{table}[htbp]
\centering
\small
\begin{adjustbox}{max width=\linewidth}
\begin{tabular}{lcccc}
\toprule
Method & EvRecall@5 & EvNDCG@5 & CrossEv@5 & EDR($\downarrow$) \\
\midrule
Naive concat (fixed order) & 0.834 & 0.542 & 0.315 & 0.485\\
Score-only rerank (no graph) & 0.846 & 0.612 & 0.392 & 0.442\\
Graph-support fusion (ours) & 0.851 & 0.625 & 0.582 & 0.365\\
Graph-retrieval + fusion (ours) & 0.872 & 0.693 & 0.645 & 0.341\\
\bottomrule
\end{tabular}
\end{adjustbox}
\caption{Evidence-level evaluation on cross-KB queries with cutoff $K{=}5$.}
\label{tab:fusion}
\end{table}
\FloatBarrier

\section{Conclusion}
\label{sec:conclusion}

We investigate RAG for Chinese Tibetan-medicine QA in a multi-KB setting where KBs differ in style, density, and evidentiary authority. Our results demonstrate that (i) DAKS improves KB routing and reduces encyclopedia dominance, and (ii) alignment graph-guided fusion further strengthens cross-KB verification at the evidence level. When combined, the full system achieves the best end-to-end cross-KB evidence coverage (CrossEv@5) on our 500-query benchmark while maintaining competitive faithfulness and citation correctness. Importantly, we show that a lightweight generator (\textsc{openPangu-Embedded-7B}) can support traceable domain QA when paired with careful routing and evidence organization, offering a practical path toward deployable Tibetan-medicine assistants.

\paragraph{Limitations.}
First, all automatic metrics are computed by a single judge model (\textsc{GLM-4.7}); although this avoids evaluator mismatch, it may introduce systematic bias, and future work should include human evaluation and multiple independent judges. Second, we evaluate only one domain and one language setting (Chinese Tibetan-medicine resources), so generalization to other medical subdomains or languages remains to be validated. Third, our alignment graph relies on entity extraction and chunk metadata; extraction errors or incomplete typing may weaken graph support and cross-KB linking. Finally, our study focuses on retrieval, fusion, and evidence packing rather than training retrieval models; improving retrievers with domain supervision or preference signals could further strengthen robustness.

\bibliographystyle{unsrt}
\bibliography{ref}

\end{document}